# Clustering by Descending to the Nearest Neighbor in the Delaunay Graph Space

Teng Qiu,   Yongjie Li*

Key Laboratory of NeuroInformation, Ministry of Education of China, School of Life Science and Technology, University of Electronic Science and Technology of China, Chengdu, China
*Corresponding author. Email: liyj@uestc.edu.cn

**Abstract**: In our previous works, we proposed a physically-inspired rule to organize the data points into an in-tree (IT) structure, in which some undesired edges are allowed to occur. By removing those undesired or redundant edges, this IT structure is divided into several separate parts, each representing one cluster. In this work, we seek to prevent the undesired edges from arising at the source. Before using the physically-inspired rule, data points are at first organized into a proximity graph which restricts each point to select the optimal directed neighbor just among its neighbors. Consequently, separated in-trees or clusters automatically arise, without redundant edges requiring to be removed.

## 1 Introduction

**Physically inspired in-tree (IT) structure**: in (*1*), we propose a physically inspired method to organize the data points into an IT structure as shown in Fig. 1. In our previous works (1-4)[1], we demonstrated its potentials in two fundamental problems of unsupervised learning: clustering and dimensionality reduction. In terms of clustering, we previously focused on how to cut those undesired or redundant edges (the edges across clusters), but now we seek to prevent the undesired edges from occurring at the source.

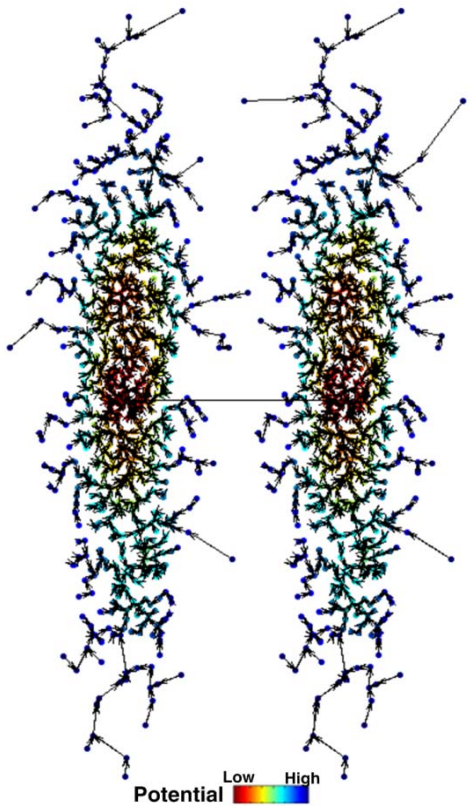

Fig. 1 An IT structure for two Gaussian datasets

## 2 Motivation

Let's reconsider the example in (*1*), where the 2-dimensional (2D) space is viewed as a horizontal rubber sheet[2]. When the data points, assumed to have mass, lie on it, the rubber sheet will curve, which, in turn, will trigger the movement of data points to the places of larger curvature (or lower potentials[3]), and eventually all points will converge at several places of the locally lowest potentials. The existence of these data points changes the feature of the rubber sheet from flat to curved, or the feature of the space from even (identical potential everywhere) to uneven (varying potentials in different places), and consequently trigger the movement of data points. Since these data points are still constrained by the space, physically speaking, it cannot happen that one point of locally lowest potential hops to other points of lower potentials. In (*1*), the physically inspired rule, basically of "descending to the nearest neighbor (DNN)", just happens to ignore the existence or the constraining role of space, and consequently the inconsistence to the physical circumstance arises (referring the redundant directed edges).

---

[1] Although the IT structure in ref. 2 is obtained based on the minimal spanning tree, but the semi-supervised cutting method there can be commonly used to cut the undesired edges in the physically inspired IT structure in ref. 1, which can lead to a more effective semi-supervised cutting method than that in ref. 1.
[2] See also http://einstein.stanford.edu/SPACETIME/spacetime2. Note that, in terms of clustering, this can be generalized to data points in high-dimensional and non-Euclidian space. Just imagine in the same manner.

Therefore, if we consider the constraining role of the space, approximate the space by a proximity graph, and similarly restrict the DNN rule in this graph space, then, the redundant (or undesired) edges can be avoided.

**3 Method**

Here, we only show the procedure (**3 steps**) of implementing this idea to the 2D data points:

**Step 1, construct Delaunay Graph.**    For the 2D dataset $\chi = \{X_i \mid i = 1, 2, ... , N\}$, the constructed[4](5) Delaunay Graph (DG), or Delaunay triangulation (6), is one of the proximity graphs, which aims to partition the space by triangular lattices, as shown in Fig. 2A.

**Step 2, identify the directed neighbor.**    First, the potential $P_i$ associated with each point $i$ is defined by

$$P_i = -\sum_{j=1}^{N} e^{-\frac{d_\chi^2(i,j)}{\sigma}}, \qquad (1)$$

where $\sigma$ is a positive parameter and $d_\chi^2(i,j)$ measures the distance between $X_i$ and $X_j$ by certain distance metric (e.g., Euclidean distance). Then, for any node $i$, its optional directed node, denoted as $I_i$, should be selected among its neighbor nodes, denoted as $N_{DG}(i)$, with lower potentials, that is, $I_i$ is the index of the point selected from the node set[5] $K_i = \{k \mid k \in N_{DG}(i), P_k < P_i\}$. If $K_i$ is not null, then $I_i$ is defined as the nearest node in $K_i$, i.e.,

$$I_i = \arg \min_{k \in K_i} d_\chi(i,k). \qquad (2)$$

---

[3] The geometric notions as "curvature", "curved space" and "the geometry of space" in General Relativity can bring convenience for us to imagine the circumstance in the 2D space, whereas bringing obstacle for even 3D cases. Therefore, in ref. 1 and this paper, we **technically** replace the notion "curvature" by a more commonly used physical notion "potential" (associated with "field") so as to make an intuitive design for our physically **inspired** "descending to the nearest neighbor"(DNN) rule. This brings to us not only the convenience to construct the "uneven" space as in Eq. 1, but also the convenience to comprehend the generalization of the DNN rule to data points to any high-dimensional Euclidean space (as the face dataset in ref. 1) and non-Euclidean space (as the mushroom dataset in ref. 1).
[4] One can use the function "Delaunay" in Matlab to obtain DG of 2D scatter plots. No parameter is involved.
[5] In practice, we have to consider the case of a group of close points, especially the overlapped ones, with the same potential, as what we've done in ref. 1, where indexes of data points are used to endow them the order.

Otherwise node $i$ has no directed node. Consequently, if we connect all the node $i$ with their corresponding end node $I_i$ (if it has), then several separated subgraphs are obtained, each being a IT structure, as shown in Fig. 2B.

**Step 3, identify the root.** The IT has several nice properties: (i) only one point has no directed node, usually called the root node; (ii) any other node has one and only one directed path to reach the root node. Therefore, the root node can be viewed as a representative of each IT structure, and by searching along the directed edges, the root node of each non-root node can be identified (Fig. 2C), or which cluster each node belongs to is identified.

## 4 Experiments

For the 2D synthetic datasets[6] (7-11) tested in our previous works, the proposed method, with an appropriate value for the parameter $\sigma$, can automatically obtain expected results for most cases (Fig. 3 A~D). This at least demonstrates the feasibility of the proposed method as an automatic clustering strategy. However, there are also failure cases (Fig. 3 E and F), which also reveals that this automatic process is not so reliable as imagined without the such needs in our previous works (1-4) as human's interaction, additional needs for labeled data, post-processing, or dimensionality reduction.

## 5 Discussions and conclusions

**Why do we use DG to approximate the 2D space?** We would like to first discuss why some other proximity graphs, e.g., $K$-Nearest-Neighbor ($K$-NN) graph, Minimal Spanning Tree (MST) (10), and Relative neighborhood Graph (RNG) (12), are inappropriate. $K$-NN brings in an additional parameter $K$. If $K$ is too small, the initial graph may under-approximate the space, whose constraining role will be exaggerated, and consequently some small yet fake clusters may occur; if $K$ is too large, the initial graph may over-approximate the space, whose constraining role will be weaken and consequently the undesired edge may still arise. Although MST and RNG share the parameter-free feature of DG, they are too sparse [in fact, they are sub-graphs of DG, as shown in ref. (12)], in effect, equivalent to $K$-NN with small $K$. In comparison, DG is just a compromise proposal.

**What about the high-dimensional Euclidean space?** Let's first consider the features of Euclidean space. It is "flat" and "continuous". Accordingly, two guidelines can be used to construct the "equivalent" graph: (**Gi**) The basic lattices of the graph should not be overlapped, namely no redundancy; (**Gii**) all lattices should fill the space (ignoring the outsider space of datasets), namely no remaining. This also gives the answer to why DG works for the 2D space, since these two guidelines are met (see Fig. 2A). Similarly, the basic lattice of the approximated graph should be a line segment for the 1D space[7], and a tetrahedron (or 3-simplex[8]) for 3D space, etc. We give these graphs a unified name, the min-max spanning graph[9] (**M²SG**), for which, "min" corresponds to Gi and "max" refers to Gii.

---
[6] Downloaded from http://cs.joensuu.fi/sipu/datasets; http://people.sissa.it/~laio/Research/Res_clustering.php
[7] The approximated graph for the 1D space turns out to be the MST.
[8] A "simplex" is the generalization of a triangle in any dimension. 1-simplex is a line segment; 2-simplex is a triangular; etc. see details in http://en.wikipedia.org/wiki/Simplex

**The problems and solutions:** In this method, the labeled data is hard to play a role, and human users can not participate in even when dealing with 2D dataset, since no intermediate results are shown to serve as a reference for them to make an effective interactive decision. The only controllable thing is the free parameter $\sigma$. However, a large $\sigma$ will result in under-partitioning of the data points, while a small $\sigma$ will lead to over-partitioning of the data points. Since no intermediate information can be used, one can only evaluate the ultimate result by some clustering evaluation indexes so as to judge the current value of $\sigma$, whereas this relies how reliable these indexes are. What's worse, failures may occur to the dataset in Fig. 3 A to D even with good parameter. To be concrete, the number of neighborhoods of each point in DG is still limited and consequently some points may become artificial root nodes (the cluster number in Fig. 3 A to D will change even with a slight change to $\sigma$). Maybe one can modify this by smoothing the distribution of the potentials in DG by Gaussian kernel, but its role should be limited.

**The meanings of this paper:** (i) it provides an automatic clustering strategy based on the physically inspired rule; (ii) it could be a new journey for this physically inspired IT structure to the field of computational geometry; (iii) as the problems stated in the above paragraph, in this work, the pursuit of an one-time solution for the problem, or a mimicking that is completely in accordance with the physical circumstance, turns out to be not so advisable in rule design. However, it conversely demonstrates the effectiveness of the philosophy underlying the design in (*1*), where a simple yet effective rule followed by a reliable repair mechanism, quite similar to the process in cell replication[10], brings both efficiency in process and reliability in result. Moreover, the design in (1) creates an attractive intermediate product as in Fig. 1, whose attractiveness is not only from the effective parts (referring that it reveals the distribution of the dataset with a sparse structure), but also from the imperfect parts (referring the undesired or redundant edges) whose salient features leave infinite space of imagination for repairing it. So, the most significant thing for this paper is not the method itself, but a negative example to demonstrate the efficiency and beauty in ref. 1, where efficiency happens to be out of mistake[11] and beauty happens to be out of redundancy[12].

---

[9] The generalization of the triangulation or "Delaunay" in d-dimensional Euclidean space meets it. See http://en.wikipedia.org/wiki/Triangulation_(geometry)

[10] Rather than an elaborate error-free process, errors are allowed to occur and then repaired by some reliable mechanisms.

[11] The "efficiency out of mistake" refers to the rule out of neglecting the role of space in ref. 1.

[12] The "beauty out of redundancy" refers to the IT structure with redundant edges in ref. 1. (Fig. 1 in this paper is one representative)

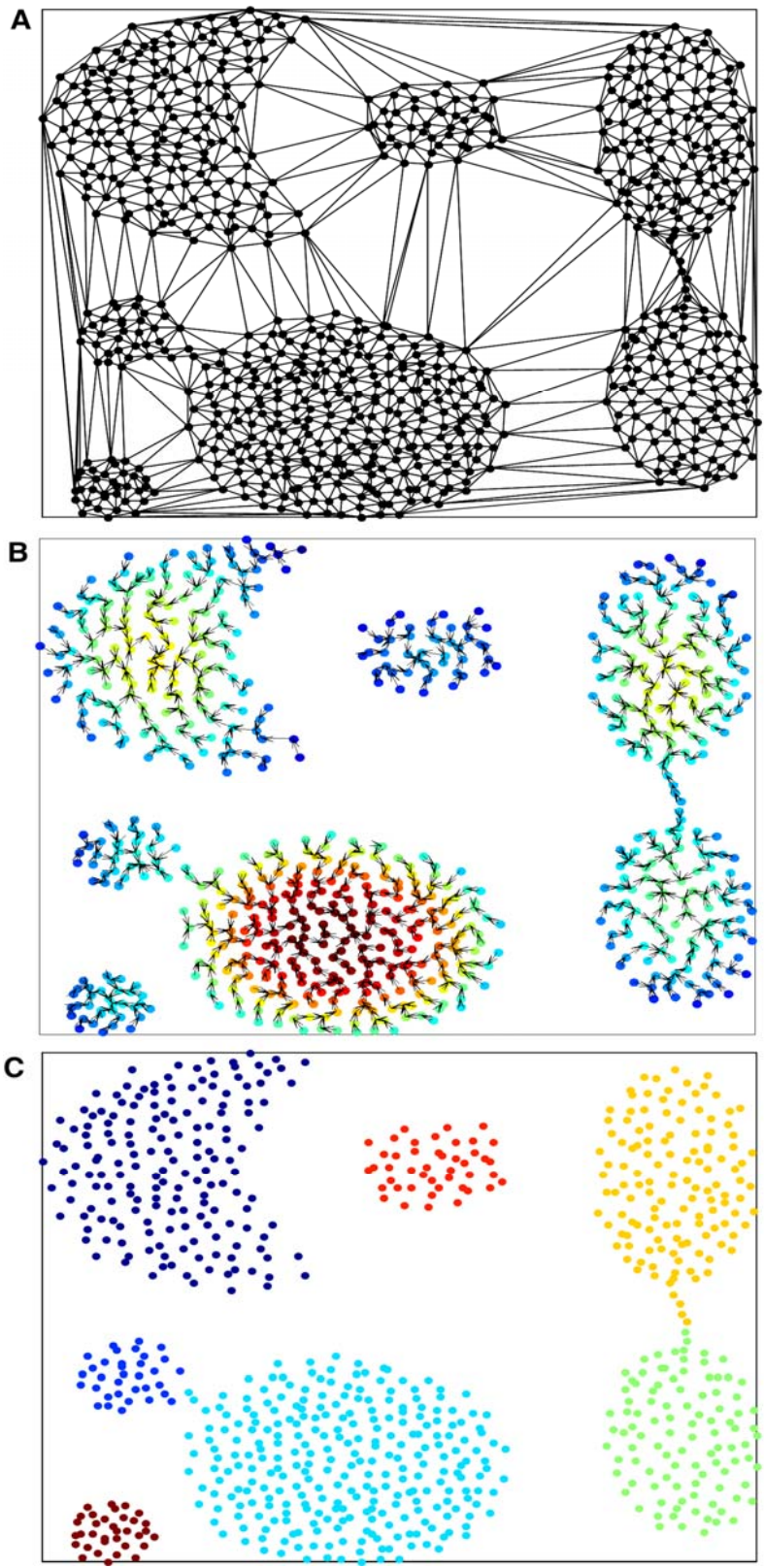

Fig. 2. An illustration of the proposed method.

(**A**) The Delaunay Graph.
(**B**) Result of step 2 ($\sigma = 2$). Colors on points denote different potential values.
(**C**) Result of step 3. Colors on points denote different clustering assignments or different roots to which points belong.

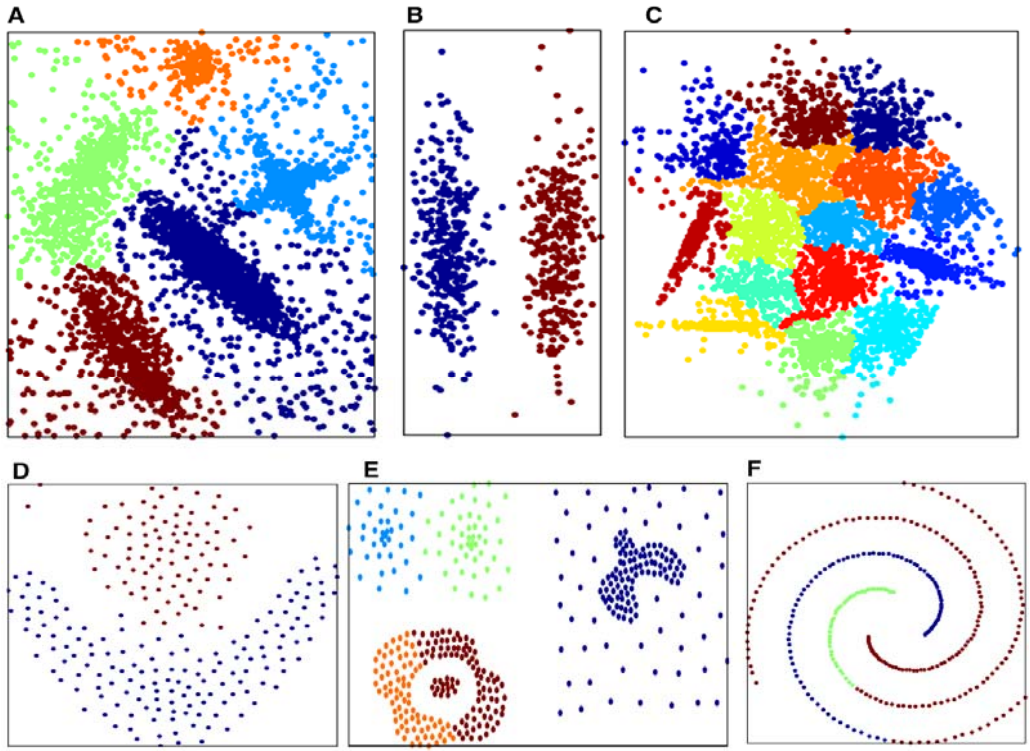

Fig. 3. Tests on several synthetic data sets. (A~D) Clustering results are quite consistent with visual perception. (E) the number of clusters is one more than that of visual perception. (F) The clustering result is totally wrong. From (A) to (F), σ = 0.05, 5, 30000, 1, 1.5, 5.

# References

1. Qiu T, Yang K, Li C, & Li Y (2014) A Physically Inspired Clustering Algorithm: to Evolve Like Particles. *arXiv preprint arXiv:1412.5902*.
2. Qiu T & Li Y (2014) An Effective Semi-supervised Divisive Clustering Algorithm. *arXiv preprint arXiv:1412.7625*.
3. Qiu T & Li Y (2015) A Generalized Affinity Propagation Clustering Algorithm for Nonspherical Cluster Discovery. *arXiv preprint arXiv:1501.04318*.
4. Qiu T & Li Y (2015) IT-map: an Effective Nonlinear Dimensionality Reduction Method for Interactive Clustering. *arXiv preprint arXiv:1501.06450*.
5. Barber CB, Dobkin DP, & Huhdanpaa H (1996) The quickhull algorithm for convex hulls. *ACM Transactions on Mathematical Software (TOMS)* 22(4):469-483.
6. Delaunay B (1934) Sur la sphere vide. *Izv. Akad. Nauk SSSR, Otdelenie Matematicheskii i Estestvennyka Nauk* 7(793-800):1-2.
7. Rodriguez A & Laio A (2014) Clustering by fast search and find of density peaks. *Science* 344(6191):1492-1496.
8. Fränti P & Virmajoki O (2006) Iterative shrinking method for clustering problems. *Pattern Recognit.* 39(5):761-775.
9. Fu L & Medico E (2007) FLAME, a novel fuzzy clustering method for the analysis of DNA microarray data. *BMC Bioinf.* 8(1):3.
10. Zahn CT (1971) Graph-theoretical methods for detecting and describing gestalt clusters. *IEEE Trans. Comput.* 100(1):68-86.
11. Chang H & Yeung D-Y (2008) Robust path-based spectral clustering. *Pattern Recognit.* 41(1):191-203.
12. Toussaint GT (1980) The relative neighbourhood graph of a finite planar set. *Pattern Recognit.* 12(4):261-268.